\documentclass[letterpaper, 10 pt, conference]{ieeeconf}  % Comment this line out if you need a4paper
\usepackage{amsfonts}
\usepackage{graphicx}
\usepackage{amsmath}
\usepackage{multirow}
\usepackage{url}
\usepackage{algorithm}
\usepackage{algpseudocode}
\usepackage{tabularx}
\usepackage{float}
\usepackage{threeparttable}
\usepackage{hyperref}

\IEEEoverridecommandlockouts                              % This command is only needed if 
\title{\LARGE \bf
$\text{(}\text{Real2Sim}\text{)}^{-1}$: 3D Branch Point Cloud Completion for Robotic Pruning in Apple Orchards
}

\author{Tian Qiu$^{1}$, Alan Zoubi$^{2}$, Nikolai Spine$^{3}$, Lailiang Cheng$^{4}$, Yu Jiang$^{5}$% <-this % stops a space
\thanks{$^{1}$Tian Qiu is with School of Electrical and Computer Engineering, Cornell University, Ithaca, USA
{\tt\small tq42@cornell.edu}}%
\thanks{$^{2}$Alan Zoubi is with Sibley School of Mechanical and Aerospace Engineering, Cornell University, Ithaca, USA}%
\thanks{$^{3}$Nikolai Spine is with College of Arts\&Sciences, Cornell University, Ithaca, USA}%
\thanks{$^{4}$Lailiang Cheng is with School of Integrative Plant Science, Cornell University, Ithaca, USA}%
\thanks{$^{5}$Yu Jiang is with School of Integrative Plant Science, Cornell University, Geneva, USA
{\tt\small yujiang@cornell.edu}}%
\thanks{\href{https://github.com/suptimq/Real-Sim_3D_Branch_Completion}{Code and Data}}%
}

\begin{document}

\maketitle
\thispagestyle{empty}
\pagestyle{empty}

%%%%%%%%%%%%%%%%%%%%%%%%%%%%%%%%%%%%%%%%%%%%%%%%%%%%%%%%%%%%%%%%%%%%%%%%%%%%%%%%
\begin{abstract}

Robotic branch pruning, a rapidly growing field addressing labor shortages in agriculture, requires detailed perception of branch geometry and topology. However, point clouds obtained in agricultural settings often lack completeness, limiting pruning accuracy. This work addressed point cloud quality via a closed-loop approach, $\text{(}\text{Real2Sim}\text{)}^{-1}$. Leveraging a Real-to-Simulation (Real2Sim) data generation pipeline, we generated simulated 3D apple trees based on realistically characterized apple tree information without manual parameterization. These 3D trees were used to train a simulation-based deep model that jointly performs point cloud completion and skeletonization on real-world partial branches, without extra real-world training. The Sim2Real qualitative results showed the model’s remarkable capability for geometry reconstruction and topology prediction. Additionally, we quantitatively evaluated the Sim2Real performance by comparing branch-level trait characterization errors using raw incomplete data and the best complete data. The Mean Absolute Error (MAE) reduced by 75\% and 8\% for branch diameter and branch angle estimation, respectively, which indicates the effectiveness of the Real2Sim data in a zero-shot generalization setting. The characterization improvements contributed to the precision and efficacy of robotic branch pruning.

\end{abstract}

%%%%%%%%%%%%%%%%%%%%%%%%%%%%%%%%%%%%%%%%%%%%%%%%%%%%%%%%%%%%%%%%%%%%%%%%%%%%%%%%
\section{INTRODUCTION}

Precision crop load management is crucial to the apple industry which holds significant economic importance in the United States and accounts for a farm value of 3.2 billion dollars in 2022 \cite{USApple}. Pruning is the key to setting an optimal crop load potential for apple trees and directly influences fruit quality, tree vitality, and overall yield. Tree crop pruning is currently done by laborious manual operations, whereas the labor shortage, especially skilled workforce, not only increases the pruning cost but also reduces the pruning performance (e.g., consistency and precision) dramatically. In contrast, robotic pruning offers a solution to these challenges by leveraging computer vision and control technologies, providing consistent, efficient, and precise pruning that enhances overall orchard productivity and sustainability.

\begin{figure}[t]
  \centering
  \includegraphics[width=\linewidth]{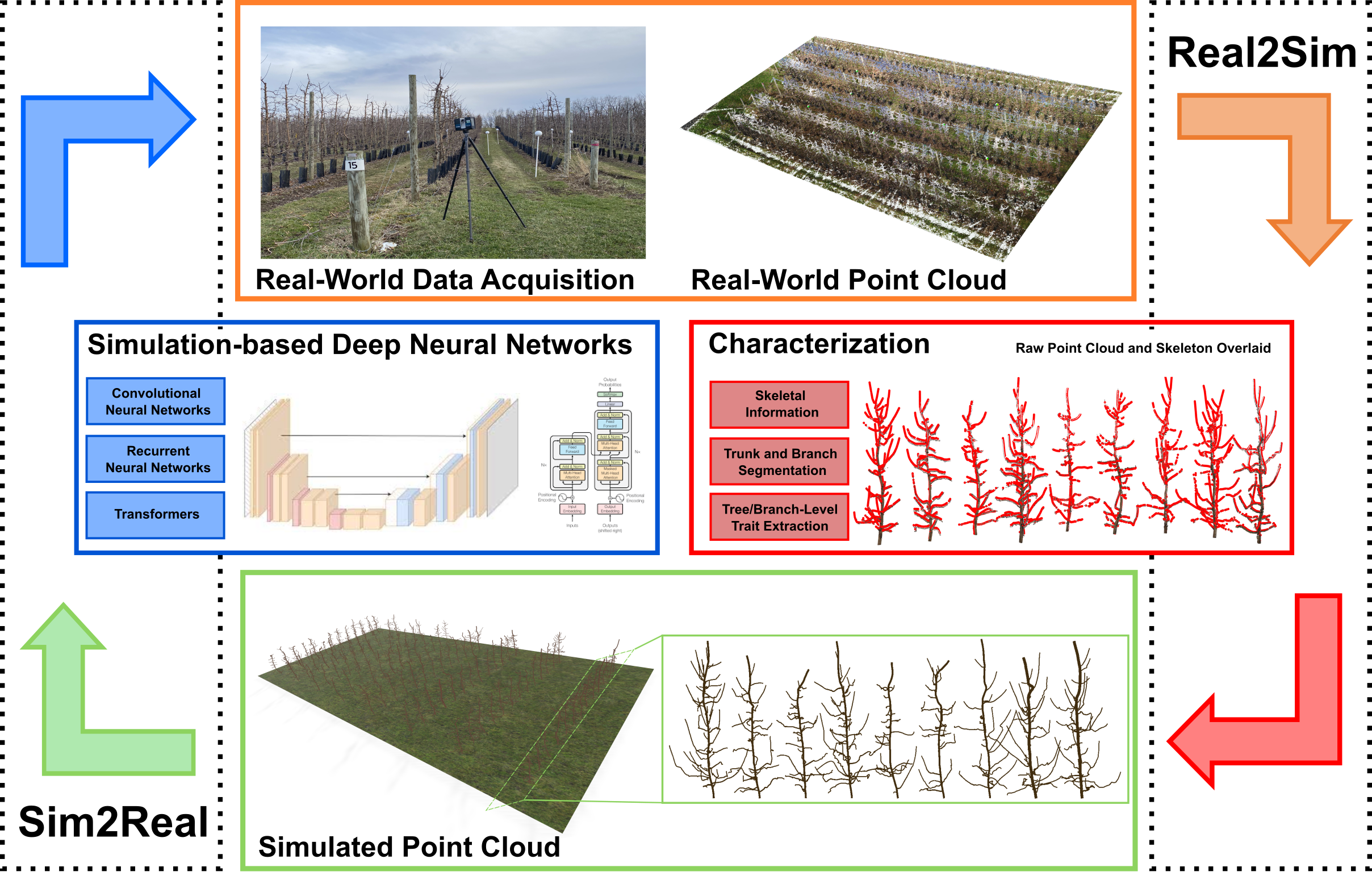}
  \caption{The proposed $\text{(}\text{Real2Sim}\text{)}^{-1}$ (i.e., Real2Sim and Sim2Real) loop for deep learning models in the context of agriculture where large-scale datasets are usually unavailable and need enormous effort for development.}
  \label{figure1}
\end{figure}

A few studies have investigated pruning robots for woody perennial crops such as apple \cite{He2018, Zahid2021}, cherry \cite{You2022}, and grape \cite{Silwal2021}. These robots were designed to use RGB-D cameras for obtaining local branch information, and subsequently prune branches or vines in a scan-and-cut manner \cite{You2020}. For instance, a pruning robot can navigate to predefined waypoints and cut a branch once a target branch is identified. While this local pruning approach has demonstrated effectiveness given simple pruning rules (e.g., removal of side branches), apple trees require a global pruning strategy to reach optimal crop load potential \cite{He2018}. This global pruning strategy considers both branch length and diameter for pruning. The length-based rule shortens branches that are longer than 40--45 cm to prevent the intervention to neighboring trees. The diameter-based rule uses both branch diameter and position (i.e., the height of a branch) to identify and remove the aged branches. Specifically, branches with a diameter exceeding 2 cm are selected as target branches. Subsequently, the largest target branch is pruned, followed by the highest one. Therefore, a global robotic pruning approach considering tree-wise branch distribution is urgently needed for apple crop load management. 

% \begin{algorithm}
% \caption{Global Branch Pruning Algorithm}\label{branch_pruning}
% \begin{algorithmic}[1]

% \Require An array $B=\{B_{1},...,B_{n}\}$ of detected branches, diameter cutoff threshold $T_{D}$, and length cutoff threshold $T_{L}$
% \Ensure An array $B_{C}$ of pruned branches and scalar $count$ indicating total pruned branches

% \State $B_{C} \gets \emptyset$, $count \gets 0$
% \State $B_{T} \gets \{\text{Diameter}(B) \geq T_{D}\}$

% \State $t_{d1}\gets \text{argmax}(\text{Diameter}(B_T))$, $B_{T} \gets B_{T} \setminus \{B_{t{d1}}\}$
% \State $t_{d2} \gets \text{argmax}(\text{Position}(B_T))$

% \State $B_{C} \gets B_{C} \cup \{B_{t_{d1}}, B_{t_{d2}}\}$, $count \gets count + 2$
% \State $B_{R} \gets B \setminus \{B_{t_{d1}}, B_{t_{d2}}\}$, $B_{L} \gets \{\text{Length}(B_{R}) \geq T_{D}\}$
% \State $B_{C} \gets B_{C} \cup B_{L}$, $count \gets count + |B_{L}|$

% \end{algorithmic}
% \end{algorithm}

A common method for the derivation of the global branch distribution is through tree quantitative structural modeling, which aims to reconstruct a quantitative structure model (QSM) of trees by capturing essential topology, geometry, and volume properties. The properties include the quantity, length, volume, angle, and size distribution of the branch. While QSM methods have shown effectiveness in various forest applications \cite{Bornand2023, Allen2023}, their application to apple tree analysis faces significant challenges. The most prominent one is the inconsistency in QSM results arising from data quality issues \cite{Qiu2024}. Point clouds of apple trees acquired under field conditions suffer from incompleteness and discontinuity because of two primary factors: 1) modern orchards have high tree density that leads to high occlusion and low visibility to sensors and 2) adverse weather conditions such as variable sunlight and strong winds introduce noise and ghost points that further exacerbate data quality issues. Overcoming these challenges requires accurately reconstructing complete point clouds from incomplete raw data, so that branch information of individual trees can be used for global robotic pruning strategy.

% Modern apple orchards have embraced the trellising training system to dictate the growing pattern of apple trees, facilitating high-density planting configurations that promise optimal profitability. This intensified planting strategy invariably results in densely populated and frequently intersecting branches, thereby impeding sensor visibility and yielding incomplete object representations within the resulting point clouds. Moreover, adverse weather conditions such as variable sunlight and strong winds introduce noise and ghost points. These undesirable points significantly contribute to point cloud incompleteness during post-data collection processes like denoising and point cloud registration. Therefore, accurately recovering complete point clouds from partial and sparse raw data is urgently needed to support the accurate characterization of apple tree traits which is crucial for a precise and efficient global robotic pruning approach.

% While realistic tree modeling has been a long-lasting research area, most of the efforts have been on naturally grown forest trees or urban trees. On the other hand, apple trees as a fruit tree and an important specialty crop have significantly different structures than forest trees and urban trees due to the well-established training systems.

In this study, we introduced the $\text{(}\text{Real2Sim}\text{)}^{-1}$ framework, a closed-loop methodology that centers around the Real2Sim data generation and Sim2Real application, aiming at providing a transformative solution for domains facing the data scarcity challenge to unlock the potential of learning-based approaches (Figure \ref{figure1}). To demonstrate the effectiveness of this loop, we applied it to address the prevalent data quality issues encountered in robotic pruning. Particularly, we utilized the Real2Sim data generation pipeline to produce simulated 3D apple tree models. These models were crafted from realistic apple tree geometric and topological data extracted by an apple tree characterization pipeline (referred to as \textbf{AppleQSM}) \cite{Qiu2024}. The simulated data was then used to train a joint completion and skeletonization model, achieving satisfactory completion results on real-world data in a zero-shot generalization setting. The Sim2Real capabilities showcased by the trained model underscored the efficacy of the closed-loop approach, both qualitatively and quantitatively, in addressing challenges in robotic pruning. Additionally, this framework opens opportunities for Digital Twins and Extended Reality by enabling real-time updates with high-fidelity simulated data, enhancing the accuracy and realism of virtual environments for and real-time control.

In summary, the major contribution of this study can be summarized in the following key points: 1) Unlike prior work that solely focused on Sim2Real, this is, to our best knowledge, the first study to propose a closed-loop approach in agriculture. This innovative method facilitates the adaptability and performance of learning-based approaches in domain areas, 2) the development of a joint transformer-based model coupled with the integration of a variance loss function, aimed at improving precision in predictions, and 3) the validation of the joint model's Sim2Real capabilities, particularly in the context of robotic branch pruning, showcasing its efficacy in real-world applications.

\section{RELATED WORK}

\subsection{3D Tree Analysis}

% TreeQSM \cite{TreeQSM} is a representative of segmentation-based QSM approaches, which entail initial segmentation of the tree point cloud into smaller subsets, followed by procedural connection to reconstruct the tree's topological structure. On the other hand, AdQSM \cite{AdQSM} is a representative of skeleton-based approaches that directly extract skeletal curves from raw input point clouds and are more robust to outliers and noise.

Analyzing trees using 3D computer vision techniques has been a long-lasting research area for forestry with a focus on tree inventory and biomass estimation \cite{Bornand2023,Allen2023}. In agriculture, tree analysis provides critical information essential to making pruning decisions \cite{Akbar2016, Tabb2017}. Current pipelines for 3D tree analysis are broadly categorized into geometry-based and learning-based approaches. A prominent tree modeling method within the geometry-based paradigm is QSM methods \cite{TreeQSM, AdQSM}. Learning-based tree analysis approaches involve deep learning models to capture complex patterns and relationships in tree structures, providing more accurate and efficient analysis \cite{Wilkes2023, Dong2023}. These methods usually use deep neural models for the segmentation of raw point cloud data and process each segmented component separately to derive traits of interest. 

% In the agricultural domain, \cite{Dong2023} developed a 3D apple tree characterization pipeline to extract individual 3D apple traits and 3D mapping for various apple training systems. They developed a 3D instance segmentation model by leveraging sparse convolutional layers and extracted fruit traits by geometry-based methods.

\subsection{3D Point Cloud Completion in Agriculture}

Previous studies on point cloud completion for agricultural applications could be categorized into template-based and learning-based approaches. Template-based methods use fitting algorithms and determine the fitting template by the prior geometric properties of objects for the prediction of missing point clouds \cite{Gene-Mola2021,Marks2022}. However, template-based approaches heavily rely on the object geometry, and their optimal performance depends on extensive parameter tuning.
In contrast, learning-based approaches leverage the strong expressiveness of deep neural networks to learn the geometric and topological features from incomplete objects and to complete missing parts. \cite{Chen2023} proposed a deep learning-based pipeline for completing point clouds of plant leaves to estimate leaf area. \cite{Xu2023} developed a deep neural network for tree completion to support robotic operations. Another line of work uses deep neural networks to learn the signed distance fields (SDF) for the completion of fruits \cite{Pan2023, Magistri2022}.

% \subsection{Simulated Tree Modeling and Simulation Deep Model}
\subsection{Sim2Real Learning-based Models for Trees}

Existing tree modeling studies fall into two categories: procedural and inverse procedural modeling. Procedural methods, like L-systems \cite{Prusinkiewicz1986,Prusinkiewicz2012}, are powerful for simulating tree geometries but require manual parameter adjustment and domain expertise, leading to time-consuming and intricate processes. Conversely, inverse procedural techniques leverage experimental data (e.g., images and range scans), to automatically estimate parameters, offering a more accessible solution that allows a broader range of users to participate without requiring intricate procedural knowledge \cite{Wang2018, Lee2023}. 

The progress in tree modeling techniques has opened doors for domains with limited large-scale real-world tree data to utilize learning-based methods \cite{You2022-IROS, You2023-RL, Bryson2023}. Nonetheless, a significant hurdle with simulation-based approaches lies in the disparity between simulated and real-world data, posing challenges for Sim2Real knowledge transfer. To address this gap, recent advances in domain adaptation techniques, as demonstrated by \cite{WangMei2018}, aimed to enhance the Sim2Real performance of simulation-based models by leveraging knowledge learned from closely related simulation source domains. Additionally, domain randomization techniques have emerged as a popular approach to mitigate the domain gap by forcing networks to learn semantically relevant features that are invariant to superficial properties \cite{James2019}. This is achieved through the introduction of variations in parameters, textures, and lighting conditions within the simulation environment. In agriculture, \cite{You2022-IROS} pioneered the first attempt to fully train a simulation-based deep model using domain randomization techniques. While their work showed promising Sim2Real branch segmentation performance across varying environments, a notable performance gap remained due to data distribution dissimilarity. Additionally, aligning semantically relevant features through domain randomization may not fully capture geometric and topological aspects.

\section{METHODOLOGY}

\subsection{Real2Sim Data Generation Pipeline}

% While current Sim2Real studies in agriculture demonstrate some effectiveness across various tasks, they often rely on manually crafted simulation scenes, leading to simulated objects that may lack realistic appearances or structures \cite{You2022-IROS, You2023-RL}. This discrepancy is primarily due to the complex biological processes of tree growth, influenced by internal signaling such as hormones and external environmental factors like gravitropism and phototropism. Moreover, the use of trellising training systems further complicates modeling for apple trees, making it difficult to ensure accurate alignment of structure and topology manually. Consequently, the data distribution disparity between simulated and real-world data inevitably restricts Sim2Real performance, especially in zero-shot generalization scenarios. To mitigate this gap, we proposed a Real2Sim data generator for producing simulated 3D apple tree models that accurately reflect the geometry and topology observed in real-world apple trees (Figure \ref{figure1}).

Current Sim2Real studies in agriculture, while effective, often rely on manually created simulation scenes, leading to unrealistic simulations \cite{You2022-IROS, You2023-RL}. This is due to the complexity of tree growth processes influenced by internal signaling and external environmental and management factors such as the use of trellising training systems. The discrepancy between simulated and real-world data limits Sim2Real performance, particularly in zero-shot generalization scenarios. To address this, we introduced a Real2Sim data generator to produce accurate 3D apple tree models.

Since tree branches can be modeled as cylinders, our process began with the structural modeling of real-world tree point clouds using AppleQSM. AppleQSM extracted skeleton points and their associated radii, forming \textbf{skeletal spheres} which were used as input for the data generator. We developed a program using the Fusion 360 API (Autodesk, version 2.0.17453) that generated 3D tree models hierarchically, starting with the trunk and then growing branches. The process involved creating a 2D spline using the coordinates of the skeletal spheres, and then performing a sweep operation to construct a 3D cylindrical body. The sweep function incorporated a taper angle to simulate branching. In this study, the taper value was empirically set to -0.5 (negative for decreasing radius). This tapering approach was chosen over directly utilizing raw skeletal spheres' radii due to observed errors in skeletal radius towards branch ends, attributed to the increasing incompleteness of the point cloud. In addition to 3D meshes, we created 100 new skeleton points by evenly sampling from the spline to serve as the ground truth.

% We employed the Fusion 360 (Autodesk, version 2.0.17453) API to develop a program. This program employed a hierarchical approach to generate 3D tree models, initiating trunk creation and subsequently growing branches from the trunk. Specifically, the trunk and branch generation process involves utilizing the coordinates of all corresponding skeletal spheres to create a 2D spline. A sweep operation was then performed, incorporating the spline and the radius of the first skeletal sphere to construct a 3D cylindrical body in a mesh representation. The sweep function also incorporates a taper angle, determining the reduction of radii for successive skeletal spheres along the body to simulate branching. In this study, the taper value was empirically set to -0.5 (negative for decreasing radius). This tapering approach was chosen over directly utilizing raw skeletal spheres' radii due to observed errors in skeletal radius towards branch ends, attributed to the increasing incompleteness of the point cloud. In addition to 3D meshes, 100 new skeleton points were created by evenly sampling from the spline to serve as the ground truth.

% In addition to 3D meshes, 100 new skeleton points were generated by evenly sampling from the spline as skeleton ground truth information.

\subsection{Joint Completion and Skeletonization Model}

% We formulated the problem of joint completion and skeletonization as estimating the completion point cloud $P_{\text{r}} \in \mathbb{R}^{K \times 3}$ and the skeleton points $P_{\text{s}} \in \mathbb{R}^{N \times 3}$, given the input partial point cloud $P_{\text{i}} \in \mathbb{R}^{K_{\text{i}} \times 3}$. The ground truth completion point cloud was denoted by $P_{\text{gt}} \in \mathbb{R}^{K \times 3}$ and the ground truth skeleton points by $P_{\text{sgt}} \in \mathbb{R}^{N \times 3}$. The joint model first produces the coarse completion point cloud $P_{\text{c}} \in \mathbb{R}^{K_{\text{c}} \times 3}$ as the intermediate representation then outputs the refined completion point cloud $P_{\text{r}}$ and the skeleton points $P_{\text{s}}$. 

\begin{figure}[htbp]
  \centering
  \includegraphics[width=\linewidth]{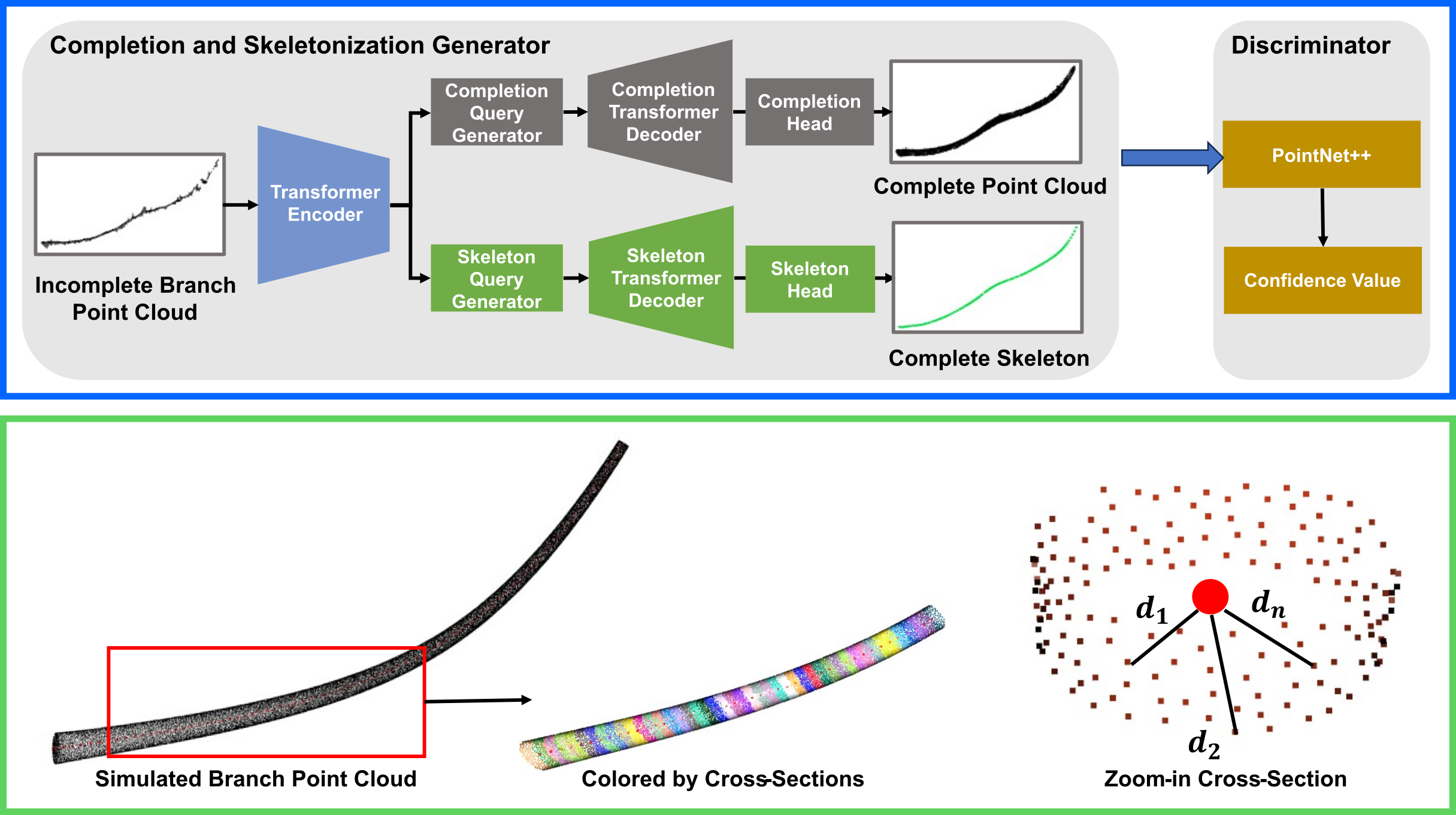}
  \caption{Top: The developed joint completion and skeletonization model formulated as a GAN framework. Bottom: The illustration of the variance loss that aims to minimize the variance of the distance from surface points ($d_{1}, d_{2}, \cdots, d_{n}$) to the skeleton point to better constrain the geometry. The red dot represents the skeleton point for the cross-section.}
  \label{figure2}
\end{figure}

The joint model consists of a transformer encoder and a joint decoder (Figure \ref{figure2}). The encoder extracts rich general features from the input partial point cloud $P_{\text{i}} \in \mathbb{R}^{K_{\text{i}} \times 3}$ and outputs point feature embedding and the coarse completion point cloud $P_{\text{c}} \in \mathbb{R}^{K_{\text{c}} \times 3}$. Employing the embedding and the coarse completion, the joint decoder generates respective geometric and topological features for heads to produce the refined completion point cloud $P_{\text{r}} \in \mathbb{R}^{K \times 3}$ and the skeleton points $P_{\text{s}} \in \mathbb{R}^{N \times 3}$. Additionally, we formulated this joint model using the generative adversarial network (GAN) to enhance the global feature learning.

\subsubsection{Transformer Encoder} 

We employed the concept of \textit{point proxy} introduced in \cite{Yu2023} to represent the input point cloud as a collection of these proxies. The transformer encoder \cite{Yu2023}, comprising geometry-aware self-attention modules, was used to capture the geometric relationships among the point proxies, ultimately generating the point query embedding $Q'$. The global features of these embeddings were used to produce the coarse completion results $P_{\text{c}}$.

\begin{equation}
\small
\label{eq:1}
\begin{aligned}
    Q' = \text{MLP}_{\text{e}}(E(P_{\text{i}})), \quad P_{\text{c}} = \text{MLP}_{\text{co}}(M(Q'))
\end{aligned}
\end{equation}

\noindent where $E$ and $M$ are the transformer encoder and the max-pooling operation, respectively.

\subsubsection{Joint Decoder} 

Point queries were generated using the global features of the point query embedding along with the coarse completion point clouds, crucially serving as the initial state for the decoder. To facilitate the smooth optimization process in the joint setting, we developed two separate query generators to produce respective point queries for the initialization of completion and skeleton decoders ($Q_{\text{c}}$ and $Q_{\text{s}}$). Employing these separate point queries, the completion decoder proceeded to generate geometric features for the completion head to produce completion results, while the skeleton decoder generated topological features for its head to output skeleton points.

\begin{equation}
\small
\label{eq:2}
\begin{aligned}
    Q_{\text{c}} = \text{MLP}_{\text{c}}([M(Q'), P_{\text{c}}]), \quad Q_{\text{s}} = \text{MLP}_{\text{s}}([M(Q'), P_{\text{c}}])
\end{aligned}
\end{equation}

% \subsubsection{Skeleton Prediction} Following \cite{Lin2021}, a skeleton point can be considered as a local center of a set of surface points. Therefore, instead of directly predicting the coordinates of the skeleton points, the convex combination of coarse completion points was used to generate the skeleton points. The weights $W \in \mathbb{R}^{K_{\text{c}} \times N}$ of the coarse completion points were predicted by MLPs and all skeleton points $P_{s}$ were derived by:

% \begin{equation}
% \label{eq:1}
% \begin{aligned}
%     P_s &= W^T P_{\text{c}} \quad \text{s.t.} j = 1, \ldots, N \quad \sum_{i=1}^{K_{\text{c}}} W(i,j) = 1
% \end{aligned}
% \end{equation}

The incorporation of the skeleton head allowed the model to acquire essential topological features, thereby improving the quality of completion results, particularly in regions with discontinuities where pure geometric features may be insufficient. Furthermore, we integrated a skeleton loss to supervise the skeleton predictions, thereby reinforcing the constraints on geometry and topology, leading to more accurate and coherent predictions.

\subsubsection{Adversarial Discriminator} 

Previous work has demonstrated that the adversarial learning strategy of the GAN framework can regularize the predictions from a global perspective and implicitly penalize outputs that deviate from the target \cite{Li2019}. Therefore, we formulated the joint model based on the GAN to enable the completion point generator to learn a richer variety of point distributions from the latent space. The discriminator aims to help implicitly evaluate the completion points produced from the generator against the latent point distribution. 

% We simply adopted the PointNet++ \cite{PointNet, PointNet++} and replaced all \textit{ReLU} layers with \textit{LeakyReLU} layers to facilitate the GAN training.

\subsection{Loss Function}

% The variance loss encourages the model to produce smooth surfaces with lower variability, enhancing the stability and consistency of the outputs.

\subsubsection{Geometry Loss} 

The geometry loss measures the difference between the predicted completion point cloud and the ground truth point cloud. Since both point clouds are unordered, the loss needs to be invariant to the permutations of the points. We adopted Chamfer Distance as the completion loss for its $O(N \log N)$ complexity \cite{Fan2017} (Equation \ref{eq:3}). Additionally, we used a repulsion loss $L_{\text{R}}$ to obtain evenly distributed points \cite{Yu2018}.

\begin{equation}
\small
\label{eq:3}
    \begin{aligned}
        L_{\text{CD}} = \frac{1}{|P_{\text{gt}}|} \sum_{p \in P_{\text{gt}}} \min_{c \in P_\text{r}} \lVert p - c \rVert + \frac{1}{|P_\text{r}|} \sum_{c \in P_\text{r}} \min_{p \in P_{\text{gt}}} \lVert c - p \rVert
    \end{aligned}
\end{equation}

\subsubsection{Variance Loss} 

Branches can be conceptualized as cylindrical objects with small cross-sections (Figure \ref{figure2}). Cylindrical objects inherently exhibit an equal-distance property, where surface points maintain a consistent distance from points along the central axis. Leveraging this geometric prior, we proposed a novel variance loss $L_{\text{V}}$ aimed at better constraining the geometry of predicted branch points. The variance loss seeks to minimize the variance of distances between ground truth skeleton points and predicted surface points (Equation \ref{eq:4}). Here, the distance from a skeleton point to a surface point is defined as the minimum distance from the skeleton point to all surface points. To ensure a close match between skeleton points and surface points, it's imperative that the completion predictions possess sound geometry. Therefore, we deferred the activation of this variance loss until the later fine-tuning stage.

\begin{equation}
\small
\label{eq:4}
    \begin{aligned}
        L_{\text{V}} &= \sum_{P=\{P_{\text{c}}, P_{\text{r}}\}} \text{Var}\left(\left\{\min_{c \in P_{\text{sgt}}} \lVert p - c \rVert_2 \mid p \in P \right\}\right)
    \end{aligned}
\end{equation}

\subsubsection{Skeleton Loss} 

We incorporated the unsupervised sampling loss \cite{Lin2021} denoted by $L_{\text{S}}$ to constrain the skeleton predictions in the first training stage. This sampling loss involves selecting points on the surface of each skeletal sphere and calculating the Chamfer Distance between these sampled points and the ground truth completion points. Furthermore, in the fine-tuning stage, we utilized the ground-truth skeleton points to supervise the joint model. This was feasible because we could easily generate ground truth data using the Real2Sim data generation pipeline. Specifically, we used the Chamfer Distance metric to compute the skeleton loss between predicted skeleton points and ground truth skeleton points $P_{\text{sgt}}$.

\subsubsection{Adversarial Loss} To supervise the discriminator, we used the least square loss as the adversarial loss:

\begin{equation}
\small
\label{eq:5}
\begin{aligned}
    L_{\text{G}} = (D_{P_{\text{gt}}}-1)^2, \quad L_{\text{D}} = (D_{P_{\text{gt}}})^2+(D_{P_{\text{r}}}-1)^2
\end{aligned}
\end{equation}

% \begin{equation}
% \label{eq:2}
% \begin{aligned}
%     L_{\text{G}} = (D_{P_{\text{gt}}}-1)^2
% \end{aligned}
% \end{equation}

% \begin{equation}
% \label{eq:3}
% \begin{aligned}
%     L_{\text{D}} = (D_{P_{\text{gt}}})^2+(D_{P_{\text{r}}}-1)^2
% \end{aligned}
% \end{equation}

\subsubsection{Joint Loss} We formulated the final joint loss (Equation \ref{eq:6}) and used $L_{\text{D}}$ to supervise the discriminator separately. We applied smaller weights to $L_{\text{S}}$ and $L_{\text{V}}$ to prioritize the learning of the completion task. It is worth noting that we used the variance loss $L_{\text{V}}$ in the finetuning stage only because the existing completion datasets do not provide ground truth skeleton points.

\begin{equation}
\small
\label{eq:6}
\begin{aligned}
    L_{J} = L_{\text{CD}}+L_{\text{R}}+L_{\text{G}}+\lambda_{1}L_{\text{S}}+\lambda_{2}L_{\text{V}}
\end{aligned}
\end{equation}

\section{Experiment}

\subsection{Dataset}

To validate the efficacy of the proposed Real2Sim concept, we developed two distinct simulated datasets, namely FB (Fusion Branch) and NB (Nozeran Branch), to train the simulation-based joint model. Additionally, we utilized a real-world dataset, referred to as COB2022, collected by a terrestrial laser scanner (TLS) in an apple orchard, to evaluate the Sim2Real performance (Table \ref{table1}).

% The first simulated dataset that was referred to as FB (Fusion Branch) dataset was generated using the Real2Sim data generation pipeline. The other simulated dataset, referred to as NB (Nozeran Branch) dataset, was generated by creating 3D branch-like objects based on the Nozeran tree growth model \cite{Tomlinson1983}.

\begin{table}[h]
\caption{Summary of datasets used in this study.}
\label{table1}
\begin{center}
\begin{tabular}{|c||c|l|l|l|}
\hline
Dataset& \#Branches&  \#Training Data&\#Testing Data&Source\\
\hline
COB2022& 106&  NA&106&TLS\\
\hline
 NB& 1432& 1145& 287&L-Py\\\hline
 FB& 1432& 1136& 296&CAD\\\hline
\end{tabular}
\end{center}
\end{table}

\subsubsection{Real-World Field Dataset} A total of 55 apple trees were used for data acquisition that was conducted during the off-season without any leaves causing occlusion to tree trunks and branches. These trees were planted at a spacing of 3.66 m (12 feet) by 0.91 m (36 inches) in 2011 and trained in the tall spindle system (Figure \ref{figure1}). We manually selected 106 branches from 9 apple trees and measured their branch diameter and angle using clipper and angle ruler, respectively, to form the COB2022 dataset. 

\subsubsection{Simulated Dataset} The rest of the 46 trees in the field dataset were used for the FB dataset generation using the Real2Sim data generation pipeline to make sure no data leakage occurs in model training. We adopted the same approach as \cite{Yuan2018} to produce point cloud data. On the other hand, we generated the NB dataset by creating randomized tree structures with varying branch cycles, axes, and radii based on the Nozeran tree growth model \cite{Tomlinson1983} in L-Py \cite{Boudon2012}. This model posits that trees grow by repeated iteration of a basic unit of structure, known as a ``tree unit''. The tree unit consists of a central leader from which lateral branches emerge following a certain pattern, which is similar to the overall apple tree structure.

\subsection{Model Training}

We leveraged the transfer learning strategy to train the simulation-based branch point cloud joint model in an end-to-end manner with a two-stage training procedure. As transformer-based models need large-scale datasets for pretraining to learn rich contextual representations and semantic relationships, we used the PCN dataset \cite{Yuan2018} to train the joint model using the joint loss $L_{\text{J}}$ without the variance loss $L_{\text{V}}$ at Stage 1. Next, we finetuned the pretrained joint model on the simulated dataset (NB or FB dataset) for efficient adaptation to domain-specific nuances by turning on variance loss $L_{\text{V}}$ at Stage 2.

The joint model was implemented using PyTorch \cite{PyTorch}. The skeleton loss weight $\lambda_{\text{1}}$ was set to 0.01 at both stages, whereas the variance loss weight $\lambda_{\text{2}}$ was set to 10 whenever it was turned on. We trained the joint model for 384 epochs with a batch size of 16 at Stage 1 and for 800 epochs with a batch size of 64 at Stage 2. More technical details for the transformer-based encoder and decoder can be found in \cite{Yu2023}.

% We utilized AdamW optimizer \cite{AdamW} with the initial learning rate as 0.0001 and weight decay as 0.0005 at both stages.
% with the continuous learning rate decay of 0.9 for every 20 epochs
% Training and validation were performed
% using a GPU server with 2 RTX A6000 GPU cards.

\subsection{Simulation-based Model Evaluation} 

We conducted the assessment to evaluate the completion performance of simulation-based models. This involved visual assessment of the completion results and quantitative comparison on COB2022 using AppleQSM that relies on geometric features of input tree point clouds to identify architectural traits. As a result, by computing the error in trait characterization for various completion models, we could accurately quantify the completion performance. It's noteworthy that while the simulation-based model achieved completion at the branch level, AppleQSM necessitated tree-level input data. Consequently, we manually segmented all primary branches for 9 apple trees in the COB2022 dataset.

% based on metrics derived from the FB dataset. Given the significant challenge of acquiring complete data in real-world scenarios, we extended our evaluation by utilizing AppleQSM to quantitatively assess the completion performance on the COB2022 dataset. 

\subsubsection{Completion Evaluation}

For the completion evaluation, we conducted a qualitative assessment of the completion results generated by various models, with a primary focus on the fidelity of geometry and topology reconstruction. As a benchmark, we employed AdaPoinTr \cite{Yu2023}, a state-of-the-art point cloud completion model, and subsequently trained four distinct models to investigate the influence of simulated datasets and model architecture. These models included AdaPoinTr-NB (AdaPoinTr trained on the NB dataset), and AdaPoinTr-FB, which aimed to underscore the efficacy of the Real2Sim dataset. In addition to it, we trained the developed joint model on the FB dataset using different loss functions to elucidate the merits of the joint design and variance loss integration. For simplicity, we referred to the joint model trained by necessary losses ($L_{\text{CD}}$, $L_{\text{R}}$, $L_{\text{G}}$, and $L_{\text{S}}$) as Joint-GS and the joint model trained by adding another variance loss as Joint-GSV.

% Joint-GS using the joint loss $L_{\text{J}}$ without the variance loss $L_{\text{V}}$, alongside Joint-GSV, trained with the full joint loss, thus elucidating the merits of the joint design and variance loss integration.

Additionally, we adopted the mean $\text{CD-}l_{\text{1}}$ (i.e., the $L\text{1}$ version of Chamfer Distance) as the metric for the quantitative evaluation of the completion performance following existing works \cite{Yu2023,Yuan2018}. We conducted the evaluation only using the FB dataset because it provides significantly more realistic geometric and topological features than the NB dataset, thereby the quantitative evaluation is more meaningful.

% The mean Chamfer Distance measures distance between the predicted completion point cloud and ground truth completion point cloud in category-level. 

\subsubsection{Characterization Evaluation}

 % Specifically, our focus lay on evaluating branch-level traits, as tree-level trait characterization was already proficient and, more importantly, branch-level traits are pivotal for robotic pruning.

% branch height (i.e., the z-axis coordinate of the first branch point)

We used the complete branch point clouds from all four models for the tree assembly, resulting in four complete tree point clouds in addition to the raw tree point cloud. These tree data were characterized by AppleQSM for the estimation of branch diameter and branch angle. By comparing the AppleQSM measurements with ground truth measurements in the COB2022 dataset, we computed metrics including mean absolute error (MAE), mean absolute percentage error (MAPE), and root mean square error (RMSE). Through these evaluations, we aimed to ascertain the efficacy and contributions of the Real2Sim dataset, the joint model, and the novel variance loss toward accurate tree characterization, which is crucial for informing robotic pruning strategies effectively.

\begin{figure}[t]
  \centering
  \includegraphics[width=\linewidth]{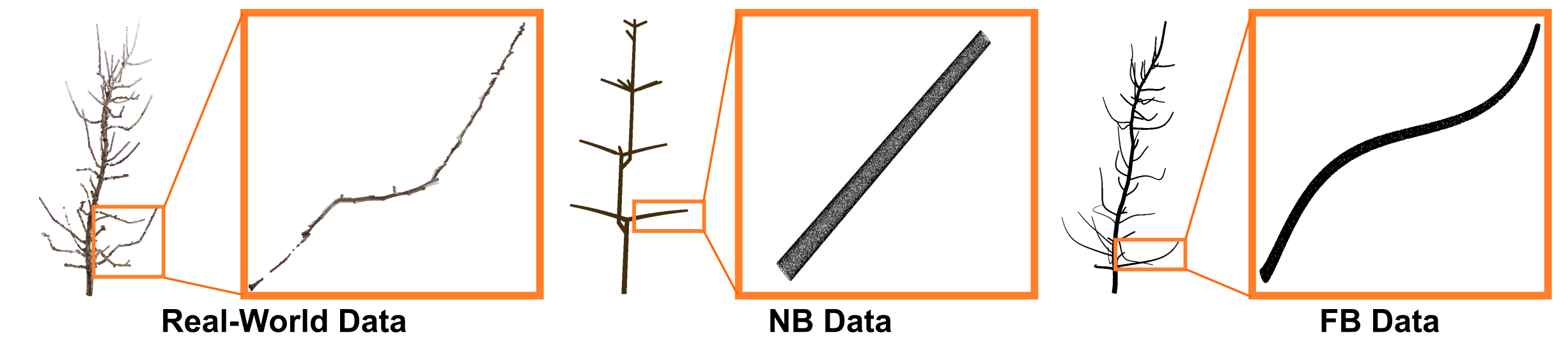}
  \caption{Branch representative from each dataset. The real-world data presents significant incompleteness and discontinuity, which is the major bottleneck for robotic perception. The NB data shows simple geometry and lacks intricate structures. The FB data shows significantly more realistic geometry and organic structure.}
  \label{figure3}
\end{figure}

\section{RESULT AND DISCUSSION}

\subsection{Real2Sim Data Generation}

The developed Real2Sim data generation presented the capability to produce a realistic representation of apple trees with intricate organic structures, indicating the effectiveness of the modeling process (Figure \ref{figure3}). Modeling trees presents a significant challenge due to their growth patterns being influenced by a mix of internal factors such as hormones, as well as external factors like gravitropism and phototropism. While existing computer graphics techniques for tree geometry generation have reached a level of maturity that allows for the creation of detailed 3D models, they often require a certain level of expertise to operate effectively. Furthermore, these techniques primarily focus on modeling forest and urban trees, which exhibit remarkably different structures compared to fruit trees such as apple trees. Notably, there is currently no publicly available dataset comprising realistic representations of apple trees suitable for training deep completion models. Through the utilization of the Real2Sim data generation pipeline, we have seamlessly bridged simulated data with real-world field data, all without the need for manual intervention, thereby ensuring the efficiency and realism of the simulation (Figure \ref{figure1}).

Furthermore, the $\text{(}\text{Real2Sim}\text{)}^{-1}$ approach holds significant promise for the agricultural robotics community, offering a transformative toolset with broad-reaching implications. By facilitating the creation of highly realistic simulations, this pipeline opens new avenues for the development and validation of learning-based robotic systems tailored specifically for various domains. The adaptability of this pipeline extends beyond apple trees, offering a versatile solution applicable to a variety of crops and tree species. This adaptability addresses the critical issue of data scarcity, providing a valuable resource for researchers working across diverse science domains.

% \begin{table}[t]
% \caption{Completion Performance Evaluation on the FB dataset.}
% \label{table2}
% \begin{center}
% \begin{tabular}{|c||c|}
% \hline
% Model& $\text{CD-}l_{\text{1}}$ ($\times$ 1000) \\
% \hline
% AdaPoinTr& 1.131 \\
% \hline
%  Joint& 1.118 \\\hline
%  Joint (Variance Loss)& 0.980 \\\hline
% \end{tabular}
% \end{center}
% \end{table}

\subsection{Sim2Real Completion Result}

\begin{figure}[htbp]
  \centering
  \includegraphics[width=\linewidth]{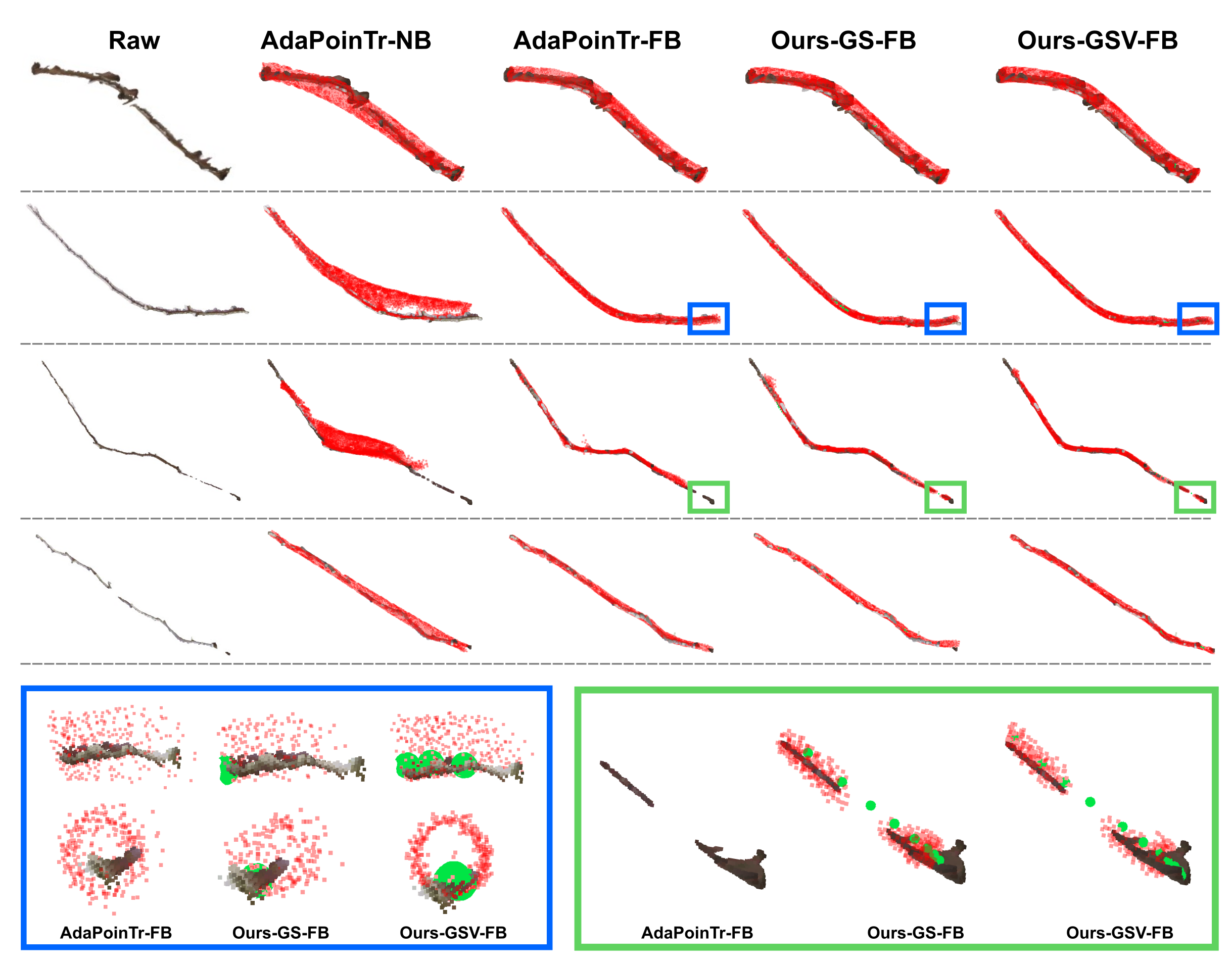}
  \caption{Completion results generated by different models. The red points represent the complete points while the green points are the skeleton points. The first two rows show the partial branches while the last two rows show the partial and discontinued branches. The bottom blue and green boxes present the zoom-in completion results of the corresponding selected area in the second and third rows. We referred to the joint model as ours. GS stands for generative and skeleton losses and V for variance loss.}
  \label{figure4}
\end{figure}

The AdaPoinTr-NB model exhibited a tendency to fit straight-up shapes for input partial branches, lacking the capability to accurately represent curvature geometry. This limitation stemmed from the inadequately realistic data representations in the NB dataset (Figure \ref{figure4}). Moreover, this model over-completed the curvature areas by adapting the straight-up geometries, which could introduce significant errors into subsequent characterization. In contrast, AdaPoinTr-FB demonstrated notably superior performance compared to AdaPoinTr-NB, offering more biologically plausible completion results. This enhancement was attributed to the improved realism provided by the FB dataset, which accurately modeled the distribution of real-world branch data and enabled the deep completion model to learn rich and realistic geometric features. However, AdaPoinTr-FB faced challenges in accurately reconstructing raw branch points around trunk-branch junction areas, crucial for precise branch-level trait extraction, particularly branch diameter estimation. Specifically, AdaPoinTr-FB provided misaligned completion results where the raw points located inside the complete points (Blue box in Figure \ref{figure4}) and completely missed one of the disconnected branch segments due to confusion caused by gaps (Green box in Figure \ref{figure4}). These completion errors constrained downstream characterization accuracy and pruning optimality.

The joint model (i.e., Joint-GS and Joint-GSV) generated better alignment and resolved disconnected segments more faithfully, implying the efficacy of the GAN framework and the joint design. The adversarial training strategy facilitates global information learning and the joint design bridges the completion head and the skeleton head thereby the skeleton predictions could provide complementary topological features for the completion. This performance improvement was confirmed by the $\text{CD-}l_{\text{1}}$ ($\times 1000$) metric where Joint-GS and Joint-GSV achieved a respective 1.118 and 0.980, outperforming the 1.131 from AdaPoinTr-FB. Furthermore, the variance loss improved the completion results qualitatively and quantitatively by encouraging more accurately filled-in points and smoother surfaces with lower variability (Figure \ref{figure4}). Without using the variance loss, the model presented relatively rough surfaces that were slightly deviated from the correct geometry due to the limitations of the Chamfer Distance loss $L_{\text{CD}}$ that could not capture all aspects of the geometric relationship given highly irregular partial branch point cloud. In contrast, the variance loss enhanced the geometric relationship by leveraging the geometric prior of cylindrical shape, providing a more coherent completion.

% \begin{table}[h]
% \caption{Branch Diameter Estimation Error from different models.}
% \label{table3}
% \begin{center}
% \begin{tabular}{|c|c|l|l|} \hline 

% Model& MAE (mm) & MAPE (\%)&RMSE (mm)\\ \hline 

% Baseline& 7.11& 47.94&7.43\\ \hline 

%  AdaPoinTr-NB& 3.43& 24.05&4.4\\ \hline 
%  AdaPoinTr-FB& 2.35& 15.22&3.62\\ \hline 
%  Joint-GS& 2.12& 14.63&2.77\\ \hline 
%  Joint-GSV& \textbf{1.77}& \textbf{12.27}&\textbf{2.28}\\ \hline
% \end{tabular}
% \end{center}
% \end{table}

% \begin{table}[h]
% \caption{Branch Angle Estimation Error from different models.}
% \label{table4}
% \begin{center}
% \begin{tabular}{|c|c|l|l|} \hline 

% Model& MAE (°)& MAPE (\%)&RMSE (°)\\ \hline 

% Baseline& 7.46& 10.11&11.22\\ \hline 

%  AdaPoinTr-NB& 9.32& 12.73&13.3\\ \hline 
%  AdaPoinTr-FB& 7.29& 10.19&9.5\\ \hline 
%  Joint-GS& 8.14& 11.17&12.31\\ \hline 
%  Joint-GSV& \textbf{6.84}& \textbf{9.42}&\textbf{9.16}\\ \hline
% \end{tabular}
% \end{center}
% \end{table}

\begin{table}[htbp]
\begin{threeparttable}
\caption{Branch-level trait estimation error from different models. We referred to the joint model as ours. GS stands for generative and skeleton Losses and V for variance loss.}
\label{table2}
\begin{center}
\begin{tabularx}{\columnwidth}{|c|c|>{\centering\arraybackslash}p{0.42cm}|>{\centering\arraybackslash}p{0.6cm}|>{\centering\arraybackslash}p{0.6cm}|>{\centering\arraybackslash}p{0.42cm}|>{\centering\arraybackslash}p{0.6cm}|>{\centering\arraybackslash}p{0.6cm}|} \hline
\multirow{2}{*}{Model} & \multirow{2}{*}{Dataset} & \multicolumn{3}{c|}{Branch Diameter} & \multicolumn{3}{c|}{Branch Angle} \\
                        &                           & MAE  & MAPE  & RMSE  & MAE & MAPE & RMSE \\ \hline
Baseline                & NA\tnote{*}               & 7.11 & 47.94 & 7.43  & 7.46 & 10.11 & 11.22 \\ \hline
AdaPoinTr               & NB                        & 3.43 & 24.05 & 4.40  & 9.32 & 12.73 & 13.30 \\ \hline
AdaPoinTr               & FB                        & 2.35 & 15.22 & 3.62  & 7.29 & 10.19 & 9.50  \\ \hline
Ours-GS                 & FB                        & 2.12 & 14.63 & 2.77  & 8.14 & 11.17 & 12.31 \\ \hline
Ours-GSV                & FB                        & \textbf{1.77} & \textbf{12.27} & \textbf{2.28} & \textbf{6.84} & \textbf{9.42} & \textbf{9.16} \\ \hline
\end{tabularx}
\end{center}
\begin{tablenotes}
\footnotesize
\item[*] The baseline model is a geometry-based model and does not require training.
\item The unit of MAE and RMSE is $mm$ and that of MAPE is \%.
\end{tablenotes}
\end{threeparttable}
\end{table}

\begin{figure*}[htbp]
  \centering
  \includegraphics[width=1.0\textwidth]{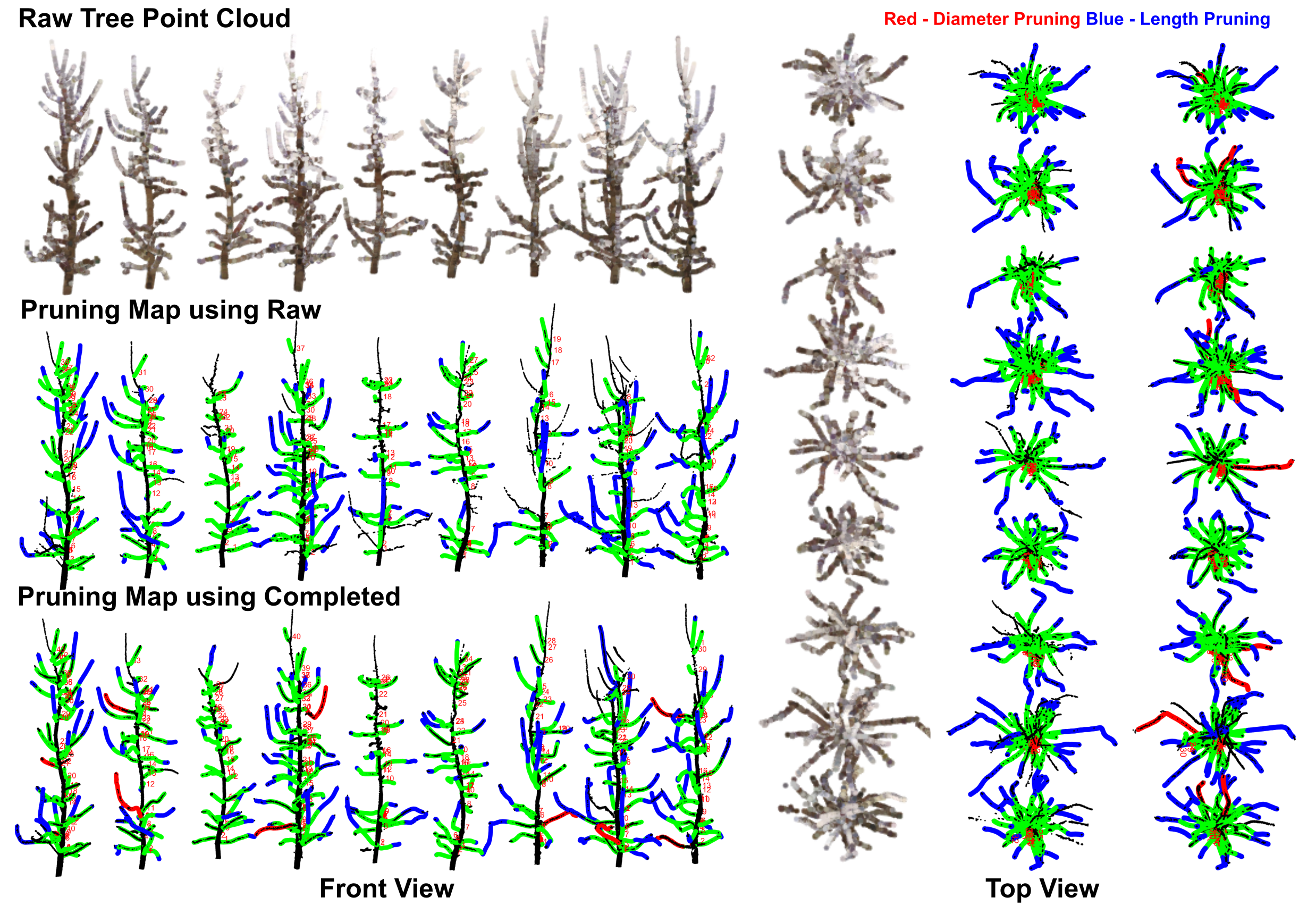}
  \caption{Pruning map developed based on the characterization results using raw data and the complete data from Ours-GSV-FB.}
  \label{figure5}
\end{figure*}

\subsection{Sim2Real Characterization Result}

The evaluation of completion models using AppleQSM revealed intriguing insights into the impact of simulated datasets and model architecture on the downstream characterization performance (Table \ref{table2}). Despite the inherent limitations of the NB dataset, derived from the Nozeran tree growth model and characterized by simplistic branch representations, AdaPoinTr-NB demonstrated a degree of efficacy, reducing the MAE of the branch diameter estimation by more than 50\%. This underscored the robustness of the baseline completion model (i.e., AdaPoinTr) and highlighted the importance of utilizing simulated datasets with high domain similarity. Conversely, the utilization of the more realistic FB dataset in AdaPoinTr-FB resulted in a substantial reduction in MAE of the branch diameter estimation, close to 70\%. This significant Sim2Real improvement underscored the crucial role played by the Real2Sim dataset in enhancing completion performance. The increased realism provided by the FB dataset enables the model to learn richer features, leading to more accurate completion results. Joint-GS demonstrated superior performance compared to AdaPoinTr-FB, leveraging the benefits of both the GAN framework and the joint setting highlighted in the completion evaluation. Building upon the success of Joint-GS, Joint-GSV achieved even greater accuracy by using the variance loss as an enhanced geometry regularization.

The characterization results obtained from Joint-GSV accurately indicated the aged branches required pruning, which was pivotal in guiding pruning robots to maintain the tree productivity in the context of apple crop load management. We generated pruning maps based on these characterization results that provide valuable insights into optimizing pruning decisions to effectively manage crop load (Figure \ref{figure5}). In contrast to raw data characterization results, which suffered from underestimation of branch diameter due to incomplete branch point cloud data, Joint-GSV significantly improved diameter estimation accuracy. This enhancement ensured that branches are accurately classified based on their position, diameter, and length, allowing for more precise identification of branches that require pruning. The importance of this precision becomes apparent when considering the outcomes of incorrect branch pruning on crop load management. Failing to accurately identify and prune branches exceeding the diameter cutoff threshold can lead to overcrowded tree canopies and inefficient resource allocation, resulting in suboptimal fruit development and reduced yield and quality. By leveraging the improved characterization accuracy provided by Joint-GSV, the pruning map offers a more nuanced understanding of the orchard’s spatial distribution. This enables targeted pruning interventions where necessary, contributing to more efficient and effective crop load management. Thus, the advancements made in this study could revolutionize the field of robotic pruning and have far-reaching impacts on agricultural practices.

\section{CONCLUSIONS}

A simulation-based deep completion network with a joint completion and skeletonization design was developed for robotic pruning in apple orchards. While the joint model was trained on simulated datasets only, its capability of generating highly accurate complete branches from real-world partial branch offered significantly enhanced geometric and topological features to enable AppleQSM to accurately characterize branch-level traits. This achievement paved the way for more accurate and efficient robotic pruning strategies. The success of the joint model's Sim2Real process owes much to the innovative Real2Sim data generation pipeline, which streamlines the creation of highly realistic apple tree models without requiring manual intervention. Moreover, we conceptualized the entire process as $\text{(}\text{Real2Sim}\text{)}^{-1}$, identifying it as a crucial component with promising potential for broader adoption across various domains. This loop addresses the challenge of domain-specific data scarcity by seamlessly bridging the gap between simulated and real-world data sources.

In future studies, we will focus on 1) training the joint model directly on tree point clouds to improve efficiency and leverage global tree information and 2) transferring the pruning map to robots for real-world pruning operations. 

\addtolength{\textheight}{-12cm}   % This command serves to balance the column lengths
                                  % on the last page of the document manually. It shortens
                                  % the textheight of the last page by a suitable amount.
                                  % This command does not take effect until the next page
                                  % so it should come on the page before the last. Make
                                  % sure that you do not shorten the textheight too much.

%%%%%%%%%%%%%%%%%%%%%%%%%%%%%%%%%%%%%%%%%%%%%%%%%%%%%%%%%%%%%%%%%%%%%%%%%%%%%%%%

%%%%%%%%%%%%%%%%%%%%%%%%%%%%%%%%%%%%%%%%%%%%%%%%%%%%%%%%%%%%%%%%%%%%%%%%%%%%%%%%

%%%%%%%%%%%%%%%%%%%%%%%%%%%%%%%%%%%%%%%%%%%%%%%%%%%%%%%%%%%%%%%%%%%%%%%%%%%%%%%%
% \section*{APPENDIX}

% Appendixes should appear before the acknowledgment.

\section*{ACKNOWLEDGMENT}

The present study was supported by the USDA NIFA Hatch project (accession No. 1025032) and USDA NIFA Specialty Crop Research Initiative (award No. 2020-51181-32197). The authors would gratefully thank Kaspar Kuehn for helping measure the branch diameter and branch angle for trees used in this study.

%%%%%%%%%%%%%%%%%%%%%%%%%%%%%%%%%%%%%%%%%%%%%%%%%%%%%%%%%%%%%%%%%%%%%%%%%%%%%%%%

% References are important to the reader; therefore, each citation must be complete and correct. If at all possible, references should be commonly available publications.

\end{document}